\pdfoutput=1
\documentclass[11pt, twocolumn]{article}
\usepackage{footnote}
\makesavenoteenv{tabular}
\makesavenoteenv{table}

\usepackage{graphicx} 

\usepackage[]{EMNLP2023}
\usepackage{graphics}
\usepackage{times}
\usepackage{latexsym}
\usepackage{titlesec}

\titlespacing*{\section}{0pt}{1ex plus 1ex minus .2ex}{1ex plus .2ex}

\usepackage[T1]{fontenc}

\usepackage[utf8]{inputenc}

\usepackage{microtype}

\usepackage{inconsolata}
\usepackage{amsmath} 

\title{Weighted Grouped Query Attention in Transformers}

\author{
  Sai Sena Chinnakonduru\textsuperscript{\textdagger}, Astarag Mohapatra\textsuperscript{\textdagger} \\
  Indiana University Bloomington \\
  \texttt{saischin@iu.edu, astmohap@iu.edu}
}

\begin{document}
\maketitle

\renewcommand{\thefootnote}{\textdagger}
\footnotetext{Equal contribution}

\begin{abstract}
The attention mechanism forms the foundational blocks for transformer language models. Recent approaches show that scaling the model achieves human-level performance. However, with increasing demands for scaling and constraints on hardware memory, the inference costs of these models remain high. To reduce the inference time, Multi-Query Attention (MQA) and Grouped-Query Attention (GQA) were proposed in \citep{shazeer2019fast} and \citep{ainslie2023gqa} respectively.

In this paper, we propose a variation of Grouped-Query Attention, termed Weighted Grouped-Query Attention (WGQA). We introduced new learnable parameters for each key and value head in the T5 decoder attention blocks, enabling the model to take a weighted average during finetuning. Our model achieves an average of 0.53\% improvement over GQA, and the performance converges to traditional Multi-head attention (MHA) with no additional overhead during inference. We evaluated the introduction of these parameters and subsequent finetuning informs the model about the grouping mechanism during training, thereby enhancing performance. Additionally, we demonstrate the scaling laws in our analysis by comparing the results between T5-small and T5-base architecture. 
\end{abstract}

\section{Introduction}

At the core of language models lies an autoregressive transformer model \citep{vaswani2023attention} that generates one token at a time based on the input sequence and the previous sequence of output tokens it has generated so far. It is a sequential process, and the workload is memory-bound \cite{kwon2023efficient}. As we scale up the model size, the inference cost becomes expensive because we need to load the model into our GPU VRAM. The original transformer paper came out in 2017 and was trained on P100 GPUs with 5.3 TFLOPs double-precision performance and 16 GB of memory, compared to the current GPU, A100, which has 80 GB of GPU memory and 9.7 TFLOPs for fp64. There has been a significant increase in the computation capability of GPUs, with only a modest increase in memory. In the ZeRO paper \cite{rajbhandari2020zero}, the authors demonstrated that GPT-2 \cite{radford2019language}, which has 1.5B parameters, required 3 GB of memory for its weights, and it could not be trained on 32 GB of memory due to the additional memory footprint of the activations and gradients. This also raises challenges in full parameter fine-tuning of these models as the memory requirements increase exponentially \cite{lv2024parameter}.

\begin{figure*}[t]
    \centering
  \includegraphics{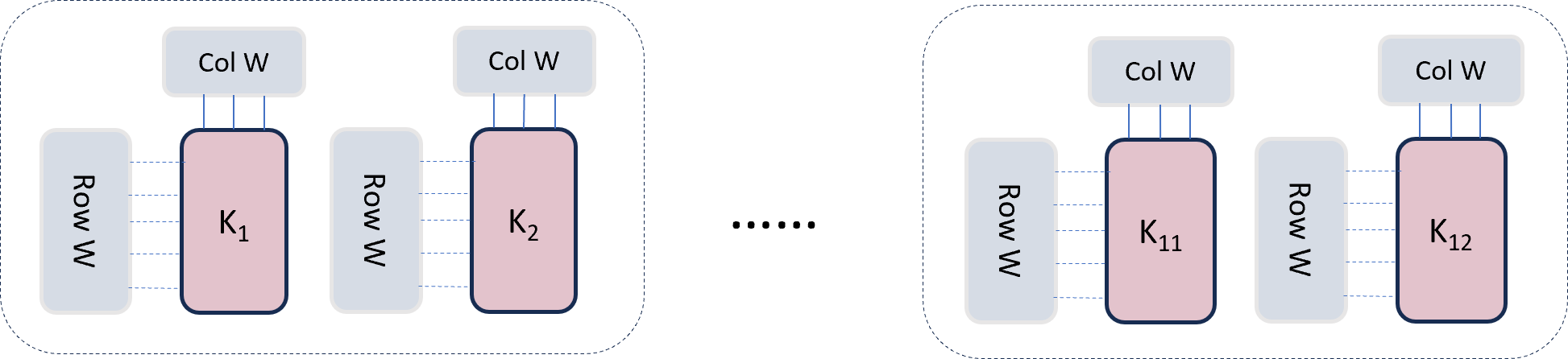}
  \caption{Grouping Key and Value heads in the decoder's attention blocks}
  \label{grouping}
\end{figure*}

The current state-of-the-art models have significantly higher parameters, which also increase the inference cost. According to a recent estimate, processing a large language model (LLM) request can be $10 \times$ more expensive than a Google search query \citealt{reutersarticle}. Due to the sequential nature of autoregressive models, the workload needs to load the model into memory and store the KV heads based on the tokens generated so far. Additionally, some decoding techniques, like beam search\cite{Freitag_2017}, can consume additional memory space by storing the KV heads for different paths and can lead to fragmentation of contiguous memory \cite{kwon2023efficient}. Hence, to resolve the memory-bound workload, the authors of the paper on MQA and GQA suggested grouping the query heads and aggregating the key-value heads after pre-training, followed by uptraining with 5-10\% of the pre-training steps and then supervised fine-tuning on a downstream task. This approach led to performance converging with MHA while being more memory efficient. In this paper, we propose a parametric way of aggregating the key-value heads (WGQA) instead of the heuristic method of taking the element-wise mean of the corresponding key and value heads. We also explore different means of aggregation to analyze whether a few additional parameters during training lead to better results. The scaling laws hold in our analysis, as the performance difference between normal GQA and our implementation widened as the parameter size increased.

\section{Related Work}
This work is focused on achieving better performance over GQA and MQA, which are similar to model-pruning methods, except that we aggregate the pruning layers. These kinds of work improve memory bandwidth and exploit the computational speed of GPUs. \cite{pope2022efficiently} showed that MQA is helpful for long input training and inference due to the reduced memory overhead.

There are other techniques for improving the memory bandwidth overhead from keys and values. Quantization \cite{dettmers2022llmint8}; \cite{frantar2023gptq} reduces the size of model parameters and activations by using INT8 or bfloat16 precision, instead of float32. There are other parameter-efficient fine-tuning (PeFT) techniques, LoRA (\cite{hu2021lora}), which decompose the projection heads into a lower dimension and then compute the gradient steps, followed by composing the full-weight matrix again for gradient update. QLoRA (\cite{dettmers2023qlora}) augmented LoRA by quantizing the static weight matrices, which further reduced the memory footprint. 

All the existing decoder-only models like Llama \cite{touvron2023llama}, Mistral \cite{jiang2023mistral}, Qwen \cite{bai2023qwen} and OLMo \cite{groeneveld2024olmo} are using grouped query attention instead of multi-head attention to reduce memory footprint. In our survey, our implementation is a novel way of grouping the key and value heads that are data-dependent and results in better performance.

\section{Method}
The attention module in the transformer architecture has three main components, query, key and value each with a dimension of $(d, d)$, where $d$ is the token embedding length. In Multi-head attention for $h$ number of heads, the projection matrices have the dimension of $(d,\frac{d}{h})$, which transforms the input embeddings $(n,d)$, where $n$ is the sequence length of the input text, to $h$ projections each of dimension $(d,\frac{d}{h})$, followed by concatenation to get the $Q$, $K$ and $V$. Then the self-attention score is given by

\begin{equation}
score = softmax\left(\frac{QK^T}{\sqrt{d}}\right) V
\label{eq1}
\end{equation}

In grouped query attention, query heads are divided into $G$ groups, reducing the number of key-value heads by a factor of $\frac{h}{G}$. Hence, the projection dimensions to obtain $Q$, $K$ and $V$ are $(n, d, d)$, $(n, d\frac{G}{h}, d\frac{G}{h})$ and $(n, d\frac{G}{h}, d\frac{G}{h})$ respectively for a batch size of 1. For GQA, $G = h/2$ and for MQA, $G = 1$. The WGQA module adds extra scalar or vector parameters depending on the configuration for key-value heads for $({w_{1,k},w_{2,k}...w_{h,k}})$ and $({w_{1,v},w_{2,v}...w_{h,v}})$. 


\begin{equation}
\small   
K = \begin{bmatrix}
\begin{pmatrix}
w_{1_k} \odot K_1 \\
+ \\
w_{2_k}\odot K_2
\end{pmatrix}
& \dots & \begin{pmatrix}
w_{(h-1)_k}\odot K_{h-1} \\
+ \\
w_{h_k}\odot K_h
\end{pmatrix}
\end{bmatrix}
\end{equation}

The modified $K$ and $V$ matrices are plugged into Eq ~\ref{eq1} for attention computation. There are additional $2h$ parameters for weighted GQA (WGQA), $2\frac{d}{h}$ (COLWGQA) for weight vectors for the columns, and $2d$ (ROWWGQA) for weight vectors for the rows in each attention layer.  These learnable parameters are multiplied with the key and value heads as shown in fig.~\ref{grouping}. The injected weights are either initialized with a value of the mean of the number of heads in a group or a random standard Gaussian distribution. This adds no additional overhead during inference, as we scale the key-value heads using learned weights after the fine-tuning process. 

\begin{table}
\noindent
\begin{tabular}{|c|c|c|c|}
\hline
\textbf{Model} & \textbf{Multi news} & \textbf{CNN} & \textbf{WMT14} \\ 
               & \textbf{R1} & \textbf{R1} & \textbf{BLEU}\\ 
\hline

\verb|MHA| & 21.7 \footnote{T5-base was not trained on multi news, hence the value is really low. The t5-large architecture achieved a 46.3 R1 score.} & 42.0 & 28 \\
\hline
\verb|GQA| & 43.5  & 41.7 & 26.1\\
\hline
\verb|WGQA| & \textbf{43.7} &  \textbf{41.9} & \textbf{26.3}\\
\hline
\verb|MQA| & 40.3 & 40.5 & 25.2\\
\hline
\verb|WMQA| & 40.7 & 40.8 & 25.5\\
\hline
\verb|ROWWGQA| & 43.6 & 41.8 & 26.0\\
\hline
\verb|COLWGQA| & \textbf{43.8} & 41.8 & 25.9\\
\hline
\verb|ROWWMQA| & 40.6 & 40.5 & 25.1\\
\hline
\verb|COLWMQA| & 40.6 & 40.7 & 25.1\\
\hline
\verb|RANDWGQA| & 42.9 & 41.9  & 25.6\\
\hline
\verb|RANDWMQA| &  37.3 & 40.7 & 25.3\\
\hline
\verb|RANDROWWGQA| & 39.7 & 40.3 & 25.2\\
\hline
\verb|RANDROWWMQA| & 36.7 & 38.9 & 23.9\\
\hline
\verb|RANDCOLWGQA| & 40.1 & 40.8  & 25.3\\
\hline
\verb|RANDCOLWMQA| & 36.5 & 39.4 & 24.4\\
\hline
\end{tabular}
\caption{Results for T5-base model with various configurations on the test set. The models prefixed with RAND signify that we initialized the weights with a random Gaussian distribution.}
\label{results}
\end{table}

\section{Implementation Details}
\subsection{Configuration}
We ran our experiments on T5-small and T5-base models implemented using Hugging Face transformers. All the models are initialized with pre-trained weights and fine-tuned on specific datasets using AdamW optimizer with 0.001 initial learning rate and scheduled linear decay. Key-value head grouping is only applied to decoder self-attention and cross-attention blocks, as mentioned in the original paper \cite{ainslie2023gqa}.

\subsection{Data and Fine-tuning}
We fine-tuned and evaluated our models using the CNN/Daily Mail, WMT 2014 German-English translation, and Multi-news datasets. We used only 500k rows for fine-tuning the WMT 2014 dataset due to limited computing resources. We trained all our models for 3 epochs with a batch size of 8 for the summarization tasks and a batch size of 32 for the translation task. We used an input length of 512 and an output length of 256 for the CNN/Daily Mail and WMT tasks. For the Multi-news summarization task, we used an input length of 2048 and an output length of 512 according to the configuration in \cite{ainslie2023gqa}. We used 4 V100 GPUs for all our experiments.

\subsection{Experimentation}
We ran all the experiments shown in table ~\ref{results} with T5-base, and with T5-small we ran only a few experiments on CNN daily mail as shown in the table~\ref{table2}.
\vspace{-5pt}
\begin{enumerate}
    \item \textbf{Weighted Grouped-Query Attention:} In this approach, new parameters, a single scalar value for each key, and a value head in the decoder's attention blocks are used. A weighted sum is then taken during the forward propagation, allowing the model to learn these parameters during fine-tuning.
    \vspace{-5pt}
    \item \textbf{Grouped-Query Attention:} In GQA, key and value heads in the decoder's attention blocks are mean pooled to form G groups \cite{ainslie2023gqa}, which are then fine-tuned.
    \vspace{-5pt}
    \item \textbf{Multi-Query Attention:} MQA involves mean pooling all key-value heads in the decoder's attention blocks to form a single key-value head that is shared across all query heads.
    \vspace{-5pt}
    \item \textbf{Weighted Multi-Query Attention:} It is similar to Weighted Grouped Query Attention, but here we just group to only one key and value head.
    \vspace{-5pt}
    \item \textbf{Row-wise Weighted Grouped-Query Attention:} Here instead of scalar weights, we introduce a column vector of size $d$ for each key and value head, which is used to scale the weights along each row as shown in fig.~\ref{grouping}.
    \vspace{-5pt}
    \item \textbf{Column wise Weighted Grouped-Query Attention:} In this, instead of scalar weights, we introduce a row vector of size $kv_{dim}$ for each key and value head, which is used to scale the weights along each column as shown in fig.~\ref{grouping}.
\end{enumerate}
For all the weighted grouped query attention configurations, we performed two types of experiments that differ in how the weights are initialized for additional introduced parameters - initializing additional parameters with weights of $kv_{heads}/h$ and random initialization. The rationale behind initializing with $kv_{heads}/h$ is that it is equivalent to starting with the mean pooled Grouped Query Attention.

\section{Results and Discussion}

\begin{figure}[hb]
    \centering
    \includegraphics[width=\columnwidth]{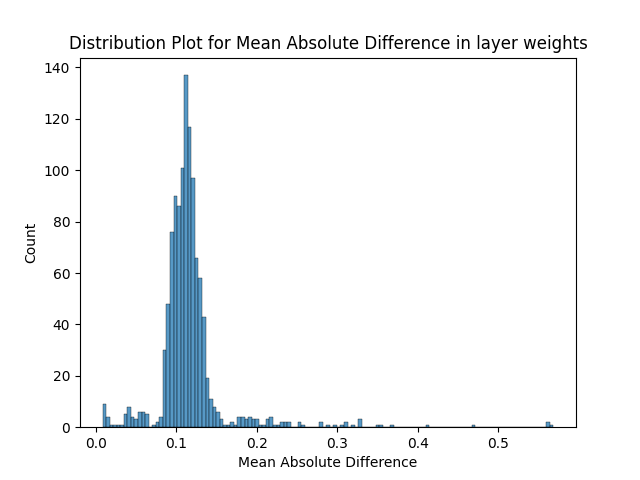}
    \caption{Distribution Plot for Mean Absolute Difference in Layer Weights}
    \label{distribution}
\end{figure}

The weighted aggregation performed better than GQA in all our experiments. The ROUGE score \cite{ganesan2018rouge} improved from 43.5 (GQA) to 43.7 (WGQA) and 43.8 (COLWGQA) for the multi-news summarization dataset. Similarly, for CNN/Daily Mail, the R1 score improved from 41.7 (GQA) to 41.9 (WGQA), and for the translation downstream task in WMT14 we reported the Bleu score \cite{Saadany_2021}, the performance improved from 26.1 (GQA) to 26.3 (WGQA) (Table~\ref{results}). During the fine-tuning stage, the number of parameters increased from GQA by 576 for WGQA, 36,864 for column-based COLWGQA, and 442,368 for row-based ROWWGQA. The WGQA performed well given the parameter and performance trade-off across the datasets.

Initializing the weights with an average of the number of heads in a group performed significantly better than random Gaussian initialization across all the datasets. Also, WMQA, which is a weighted version of MQA, performed better than MQA and approached the performance of GQA. This can lead to even more parameter savings. We validated our results with the scaling laws by testing our models on a smaller architecture, T5-small, for the CNN/Daily Mail dataset (Table~\ref{table2}). Hence, increasing the model size results in better evaluation metrics, and we believe that bigger models would widen the performance gap between WGQA and GQA.
\begin{table}
\centering
\begin{tabular}{|c|c|c|}
\hline
\textbf{MHA} & \textbf{GQA} & \textbf{WGQA} \\
\hline
 41.1& 40.3 & 40.3 \\
\hline
\end{tabular}
\caption{Rouge 1 score for CNN Daily Mail dataset of t5-small architecture}
\label{table2}
\end{table}

To check whether the learned weights in the WGQA configuration differ from those in the GQA configuration, we conducted a statistical analysis. We grouped the key and value heads of the WGQA model according to the learned weights and calculated the mean absolute loss for each layer. In the attention blocks, we calculated the mean for each head separately and observed that the weights are significantly different, with the mean absolute difference centering around 0.1 as shown in fig.~\ref{distribution}. The p-value, $1e-6$ was less than the significance level of 0.05, rejecting the null hypothesis of zero mean absolute difference.

\section{Conclusion}
This paper focuses on improving the GQA algorithm by introducing a novel way of aggregating the KV heads. From the scaling laws, we can extrapolate that the performance will improve with model size, and the models converge into different parameter spaces, as shown in the mean absolute plot. Given the prevalence of the GQA-based decoder model in Large Language Models, this technique can aid in building more accurate models with the overhead of linearly scaling weights during training only.

\section{Limitations and Future Work}
For summarization tasks, we used the ROUGE score, which is not an ideal metric and it doesn't give the whole picture to validate our increase in performance. Due to limited computing resources, we didn't pre-train our model from scratch or fine-tune on larger datasets and models, which would give better results for comparison.

In GQA, the grouped key value heads are repeated to match the dimension of query heads. In the future, we can introduce parameters that can dynamically repeat the key value heads. Specifically, in Grouped Query models such as Llama\cite{touvron2023llama} and OpenELM \cite{mehta2024openelm}, instead of sharing the key and value heads, we propose multiplying them with weights to create distinct heads. This approach would allow the model to differentiate between the heads, potentially enhancing performance. Additionally, we aim to implement this using decoder-only models, which is the current norm in language models.
\vspace{50pt}

\bibliography{anthology,custom}

\begin{thebibliography}{21}
\expandafter\ifx\csname natexlab\endcsname\relax\def\natexlab#1{#1}\fi

\bibitem[{Ainslie et~al.(2023)Ainslie, Lee-Thorp, de~Jong, Zemlyanskiy, Lebrón, and Sanghai}]{ainslie2023gqa}
Joshua Ainslie, James Lee-Thorp, Michiel de~Jong, Yury Zemlyanskiy, Federico Lebrón, and Sumit Sanghai. 2023.
\newblock \href {http://arxiv.org/abs/2305.13245} {Gqa: Training generalized multi-query transformer models from multi-head checkpoints}.

\bibitem[{Bai et~al.(2023)Bai, Bai, Chu, Cui, Dang, Deng, Fan, Ge, Han, Huang, Hui, Ji, Li, Lin, Lin, Liu, Liu, Lu, Lu, Ma, Men, Ren, Ren, Tan, Tan, Tu, Wang, Wang, Wang, Wu, Xu, Xu, Yang, Yang, Yang, Yang, Yao, Yu, Yuan, Yuan, Zhang, Zhang, Zhang, Zhang, Zhou, Zhou, Zhou, and Zhu}]{bai2023qwen}
Jinze Bai, Shuai Bai, Yunfei Chu, Zeyu Cui, Kai Dang, Xiaodong Deng, Yang Fan, Wenbin Ge, Yu~Han, Fei Huang, Binyuan Hui, Luo Ji, Mei Li, Junyang Lin, Runji Lin, Dayiheng Liu, Gao Liu, Chengqiang Lu, Keming Lu, Jianxin Ma, Rui Men, Xingzhang Ren, Xuancheng Ren, Chuanqi Tan, Sinan Tan, Jianhong Tu, Peng Wang, Shijie Wang, Wei Wang, Shengguang Wu, Benfeng Xu, Jin Xu, An~Yang, Hao Yang, Jian Yang, Shusheng Yang, Yang Yao, Bowen Yu, Hongyi Yuan, Zheng Yuan, Jianwei Zhang, Xingxuan Zhang, Yichang Zhang, Zhenru Zhang, Chang Zhou, Jingren Zhou, Xiaohuan Zhou, and Tianhang Zhu. 2023.
\newblock \href {http://arxiv.org/abs/2309.16609} {Qwen technical report}.

\bibitem[{Dastin(2023)}]{reutersarticle}
Jeffrey Dastin. 2023.
\newblock Focus: For tech giants, ai like bing and bard poses billion dollar search problem.

\bibitem[{Dettmers et~al.(2022)Dettmers, Lewis, Belkada, and Zettlemoyer}]{dettmers2022llmint8}
Tim Dettmers, Mike Lewis, Younes Belkada, and Luke Zettlemoyer. 2022.
\newblock \href {http://arxiv.org/abs/2208.07339} {Llm.int8(): 8-bit matrix multiplication for transformers at scale}.

\bibitem[{Dettmers et~al.(2023)Dettmers, Pagnoni, Holtzman, and Zettlemoyer}]{dettmers2023qlora}
Tim Dettmers, Artidoro Pagnoni, Ari Holtzman, and Luke Zettlemoyer. 2023.
\newblock \href {http://arxiv.org/abs/2305.14314} {Qlora: Efficient finetuning of quantized llms}.

\bibitem[{Frantar et~al.(2023)Frantar, Ashkboos, Hoefler, and Alistarh}]{frantar2023gptq}
Elias Frantar, Saleh Ashkboos, Torsten Hoefler, and Dan Alistarh. 2023.
\newblock \href {http://arxiv.org/abs/2210.17323} {Gptq: Accurate post-training quantization for generative pre-trained transformers}.

\bibitem[{Freitag and Al-Onaizan(2017)}]{Freitag_2017}
Markus Freitag and Yaser Al-Onaizan. 2017.
\newblock \href {https://doi.org/10.18653/v1/w17-3207} {Beam search strategies for neural machine translation}.
\newblock In \emph{Proceedings of the First Workshop on Neural Machine Translation}. Association for Computational Linguistics.

\bibitem[{Ganesan(2018)}]{ganesan2018rouge}
Kavita Ganesan. 2018.
\newblock \href {http://arxiv.org/abs/1803.01937} {Rouge 2.0: Updated and improved measures for evaluation of summarization tasks}.

\bibitem[{Groeneveld et~al.(2024)Groeneveld, Beltagy, Walsh, Bhagia, Kinney, Tafjord, Jha, Ivison, Magnusson, Wang, Arora, Atkinson, Authur, Chandu, Cohan, Dumas, Elazar, Gu, Hessel, Khot, Merrill, Morrison, Muennighoff, Naik, Nam, Peters, Pyatkin, Ravichander, Schwenk, Shah, Smith, Strubell, Subramani, Wortsman, Dasigi, Lambert, Richardson, Zettlemoyer, Dodge, Lo, Soldaini, Smith, and Hajishirzi}]{groeneveld2024olmo}
Dirk Groeneveld, Iz~Beltagy, Pete Walsh, Akshita Bhagia, Rodney Kinney, Oyvind Tafjord, Ananya~Harsh Jha, Hamish Ivison, Ian Magnusson, Yizhong Wang, Shane Arora, David Atkinson, Russell Authur, Khyathi~Raghavi Chandu, Arman Cohan, Jennifer Dumas, Yanai Elazar, Yuling Gu, Jack Hessel, Tushar Khot, William Merrill, Jacob Morrison, Niklas Muennighoff, Aakanksha Naik, Crystal Nam, Matthew~E. Peters, Valentina Pyatkin, Abhilasha Ravichander, Dustin Schwenk, Saurabh Shah, Will Smith, Emma Strubell, Nishant Subramani, Mitchell Wortsman, Pradeep Dasigi, Nathan Lambert, Kyle Richardson, Luke Zettlemoyer, Jesse Dodge, Kyle Lo, Luca Soldaini, Noah~A. Smith, and Hannaneh Hajishirzi. 2024.
\newblock \href {http://arxiv.org/abs/2402.00838} {Olmo: Accelerating the science of language models}.

\bibitem[{Hu et~al.(2021)Hu, Shen, Wallis, Allen-Zhu, Li, Wang, Wang, and Chen}]{hu2021lora}
Edward~J. Hu, Yelong Shen, Phillip Wallis, Zeyuan Allen-Zhu, Yuanzhi Li, Shean Wang, Lu~Wang, and Weizhu Chen. 2021.
\newblock \href {http://arxiv.org/abs/2106.09685} {Lora: Low-rank adaptation of large language models}.

\bibitem[{Jiang et~al.(2023)Jiang, Sablayrolles, Mensch, Bamford, Chaplot, de~las Casas, Bressand, Lengyel, Lample, Saulnier, Lavaud, Lachaux, Stock, Scao, Lavril, Wang, Lacroix, and Sayed}]{jiang2023mistral}
Albert~Q. Jiang, Alexandre Sablayrolles, Arthur Mensch, Chris Bamford, Devendra~Singh Chaplot, Diego de~las Casas, Florian Bressand, Gianna Lengyel, Guillaume Lample, Lucile Saulnier, Lélio~Renard Lavaud, Marie-Anne Lachaux, Pierre Stock, Teven~Le Scao, Thibaut Lavril, Thomas Wang, Timothée Lacroix, and William~El Sayed. 2023.
\newblock \href {http://arxiv.org/abs/2310.06825} {Mistral 7b}.

\bibitem[{Kwon et~al.(2023)Kwon, Li, Zhuang, Sheng, Zheng, Yu, Gonzalez, Zhang, and Stoica}]{kwon2023efficient}
Woosuk Kwon, Zhuohan Li, Siyuan Zhuang, Ying Sheng, Lianmin Zheng, Cody~Hao Yu, Joseph~E. Gonzalez, Hao Zhang, and Ion Stoica. 2023.
\newblock \href {http://arxiv.org/abs/2309.06180} {Efficient memory management for large language model serving with pagedattention}.

\bibitem[{Lv et~al.(2024)Lv, Yang, Liu, Gao, Guo, and Qiu}]{lv2024parameter}
Kai Lv, Yuqing Yang, Tengxiao Liu, Qinghui Gao, Qipeng Guo, and Xipeng Qiu. 2024.
\newblock \href {http://arxiv.org/abs/2306.09782} {Full parameter fine-tuning for large language models with limited resources}.

\bibitem[{Mehta et~al.(2024)Mehta, Sekhavat, Cao, Horton, Jin, Sun, Mirzadeh, Najibi, Belenko, Zatloukal, and Rastegari}]{mehta2024openelm}
Sachin Mehta, Mohammad~Hossein Sekhavat, Qingqing Cao, Maxwell Horton, Yanzi Jin, Chenfan Sun, Iman Mirzadeh, Mahyar Najibi, Dmitry Belenko, Peter Zatloukal, and Mohammad Rastegari. 2024.
\newblock \href {http://arxiv.org/abs/2404.14619} {Openelm: An efficient language model family with open training and inference framework}.

\bibitem[{Pope et~al.(2022)Pope, Douglas, Chowdhery, Devlin, Bradbury, Levskaya, Heek, Xiao, Agrawal, and Dean}]{pope2022efficiently}
Reiner Pope, Sholto Douglas, Aakanksha Chowdhery, Jacob Devlin, James Bradbury, Anselm Levskaya, Jonathan Heek, Kefan Xiao, Shivani Agrawal, and Jeff Dean. 2022.
\newblock \href {http://arxiv.org/abs/2211.05102} {Efficiently scaling transformer inference}.

\bibitem[{Radford et~al.(2019)Radford, Wu, Child, Luan, Amodei, Sutskever et~al.}]{radford2019language}
Alec Radford, Jeffrey Wu, Rewon Child, David Luan, Dario Amodei, Ilya Sutskever, et~al. 2019.
\newblock Language models are unsupervised multitask learners.
\newblock \emph{OpenAI blog}, 1(8):9.

\bibitem[{Rajbhandari et~al.(2020)Rajbhandari, Rasley, Ruwase, and He}]{rajbhandari2020zero}
Samyam Rajbhandari, Jeff Rasley, Olatunji Ruwase, and Yuxiong He. 2020.
\newblock \href {http://arxiv.org/abs/1910.02054} {Zero: Memory optimizations toward training trillion parameter models}.

\bibitem[{Saadany and Orăsan(2021)}]{Saadany_2021}
Hadeel Saadany and Constantin Orăsan. 2021.
\newblock \href {https://doi.org/10.26615/978-954-452-071-7_006} {Bleu, meteor, bertscore: Evaluation of metrics performance in assessing critical translation errors in sentiment-oriented text}.
\newblock In \emph{Proceedings of the Translation and Interpreting Technology Online Conference TRITON 2021}, TRITON 2021. INCOMA Ltd. Shoumen, BULGARIA.

\bibitem[{Shazeer(2019)}]{shazeer2019fast}
Noam Shazeer. 2019.
\newblock \href {http://arxiv.org/abs/1911.02150} {Fast transformer decoding: One write-head is all you need}.

\bibitem[{Touvron et~al.(2023)Touvron, Lavril, Izacard, Martinet, Lachaux, Lacroix, Rozière, Goyal, Hambro, Azhar, Rodriguez, Joulin, Grave, and Lample}]{touvron2023llama}
Hugo Touvron, Thibaut Lavril, Gautier Izacard, Xavier Martinet, Marie-Anne Lachaux, Timothée Lacroix, Baptiste Rozière, Naman Goyal, Eric Hambro, Faisal Azhar, Aurelien Rodriguez, Armand Joulin, Edouard Grave, and Guillaume Lample. 2023.
\newblock \href {http://arxiv.org/abs/2302.13971} {Llama: Open and efficient foundation language models}.

\bibitem[{Vaswani et~al.(2023)Vaswani, Shazeer, Parmar, Uszkoreit, Jones, Gomez, Kaiser, and Polosukhin}]{vaswani2023attention}
Ashish Vaswani, Noam Shazeer, Niki Parmar, Jakob Uszkoreit, Llion Jones, Aidan~N. Gomez, Lukasz Kaiser, and Illia Polosukhin. 2023.
\newblock \href {http://arxiv.org/abs/1706.03762} {Attention is all you need}.

\end{thebibliography}
\bibliographystyle{acl_natbib}


\end{document}